\documentclass[nouppercase]{ifmbe}
\usepackage{amsmath} 
\usepackage{textcomp} 
\usepackage{url}  
\usepackage{amsthm}
\usepackage{amssymb}
\usepackage{xcolor}
\usepackage{amsfonts}
\usepackage{caption}
\usepackage{subcaption}
\usepackage{float}
\usepackage{cite}
\usepackage[symbol]{footmisc}

\definecolor{megreen}{rgb}{0.13, 0.55, 0.13}

\title{Human Motion Detection Using Sharpened Dimensionality Reduction and Clustering}

\affiliation{Bernoulli Institute/Department of Computer Science, University of Groningen, Groningen, The Netherlands}{1}

\author{Jeewon Heo$^\dagger$}{1}
\author{Youngjoo Kim$^{*\dagger}$}{1}
\author{Jos B.T.M. Roerdink}{1}

\begin{document}

\maketitle

\begin{abstract}
Sharpened dimensionality reduction (SDR), which belongs to the class of multidimensional projection techniques, has recently been introduced to tackle the challenges in the exploratory and visual analysis of high-dimensional data. SDR has been applied to various real-world datasets, such as human activity sensory data and astronomical datasets. However, manually labeling the samples from the generated projection are expensive. To address this problem, we propose here to use clustering methods such as \emph{k}-means, Hierarchical Clustering, Density-Based Spatial Clustering of Applications with Noise (DBSCAN), and Spectral Clustering to easily label the 2D projections of high-dimensional data. We test our pipeline of SDR and the clustering methods on a range of synthetic and real-world datasets, including two different public human activity datasets extracted from smartphone accelerometer or gyroscope recordings of various movements. We apply clustering to assess the visual cluster separation of SDR, both qualitatively and quantitatively. We conclude that clustering SDR results yields better labeling results than clustering plain DR, and that \emph{k}-means is the recommended clustering method for SDR in terms of clustering accuracy, ease-of-use, and computational scalability.

\end{abstract}
\footnotetext{$^\dagger$ These authors contributed equally.}
\footnotetext{$^*$ Corresponding author (lyoungjookiml@gmail.com)}

\section{Introduction}
\label{sec:intro}
Dimensionality reduction (DR) is a commonly used approach to visualize and explore multidimensional data. Recently, Sharpened DR (SDR), a pre-processing method that enhances the cluster separation of DR, was proposed\,\cite{sdr}. The preconditioning step sharpens the sample density using Gradient Clustering (GC) in $n$-dimensional space, where the sharpening effect is visible after DR. This separability of clusters allows end-users to explore the multidimensional data more easily.

Although SDR is capable of preserving the underlying cluster structures in $n$-dimensional space and representing them in 2D, labeling these structures in the resulting projections is time consuming. Moreover, identifying the visual clusters can be dependent on subjective perception. To address this problem, we focus here on suggesting a set of clustering methods and validation metrics that will automatically label the clusters from SDR and assess their quality both qualitatively and quantitatively. In this pipeline consisting of SDR and clustering, the labeling process remains as a black box to end-users, providing a convenient integration with further steps of data exploration or analysis using SDR.

The paper is structured as follows. Sec.~\ref{sec:method}\hspace{0.3cm} explains the proposed pipeline. Sec.~\ref{sec:results}\hspace{0.3cm} shows the experimental results and Sec.~\ref{sec:discussion}\hspace{0.3cm} discusses the results. Sec.~\ref{sec:conclusion}\hspace{0.3cm} concludes the paper.

\section{Method}
\label{sec:method}
Let $D = \{\mathbf{x}_1,\ldots,\mathbf{x}_N\}$ be a set of $N$ $n$-dimensional observations, where $\mathbf{x}_i = [x^1_i$ $x^2_i$ $\cdots$ $x^n_i] \in \mathbb{R}^n$ and $x^j_i$ is the $i^{th}$ observation of the $j^{\text{th}}$ ($1 \leq j \leq n$) dimension. The sharpening method can be seen as a function $S: \mathbb{R}^{n} \rightarrow \mathbb{R}^{n}$, whereas DR is a function $F: \mathbb{R}^{n} \rightarrow \mathbb{R}^{n'}$, where $n'<<n$, where commonly $n'=2$. We set $F$ to Landmark Multidimensional Scaling (LMDS) because it shows a clear separation of clusters for human activity and motion data\,\cite{sdr}. We aim here to replicate the results from Kim \emph{et al.} to compare the \emph{clustering} results between DR and SDR, which has not been shown before\,\cite{sdr}. We also define $L$ as the list of labels for all points acquired from clustering and $G$ as the prior, which is the list of ground-truth labels of a dataset. Note that we use the terms `points', `observations', and `samples' interchangeably.


\subsection{Clustering methods}
\label{sec:method:clustering}
Clustering methods are chosen to cover different types of methods such as partitional, hierarchical, density-based, and graph-based. A popular partitional algorithm, \emph{k}-means clustering is a Euclidean distance-based algorithm, which makes it isotropic. Due to this property, it produces spherical clusters even when the actual clusters are non-spherical\,\cite{reddy}. To address this problem, density-based clustering methods, which do not make assumptions on distribution or shapes of data, have been proposed\,\cite{reddy}, thus used in this paper. Refer to Sec.~\ref{sec:discussion}\hspace{0.3cm} for more discussion on the selection of clustering methods.


\noindent\textbf{\emph{k}-means} assign points to clusters based on their minimum distance to cluster centroids, where the number of cluster centroids is equal to \emph{k}\,\cite{reddy, kmeans}. The parameter \verb!Replicates! is used to limit the maximum number of iterations during expectation-maximization and is set to 10. Distance is measured using squared Euclidean distance.

\noindent\textbf{Hierarchical clustering (HC)} varies based on the proximity measures used. We choose here two of the most used measures--\emph{complete} and \emph{ward} linkages \cite{HC:complete, HC:ward}. HC first constructs a dissimilarity matrix using one of these linkages and merges the closest clusters until all points are in a single maximal cluster\,\cite{reddy, HC:complete, HC:ward}. We build an agglomerative hierarchy using Euclidean distance.

%

\noindent\textbf{Density-based spatial clustering of applications with noise (DBSCAN)} groups closely packed points together and marks points in low-density regions as outliers or noise. Clusters are created based on core points that are selected based on the neighborhood with radius $\epsilon$ containing at least $MinPts$ points, and other points get assigned to one of these clusters\,\cite{reddy, dbscan:original}. Parameter $MinPts$ is set to $\log(N)$ and $\epsilon$ is set to the value of a point that is farthest away from a line created by connecting the first and the last points in the \emph{k}-nearest neighbors distance plot, where $k=MinPts$. Distance is measured using squared Euclidean distance.


\noindent\textbf{Spectral clustering (SC)} uses eigenvalues of graph Laplacian matrices based on the edges in a graph to cluster similar nodes. A similarity graph is constructed using \emph{k}-nearest neighbors where \emph{k} is set to $log(N)$.

\subsection{Evaluation metrics}
\label{sec:method:eval}
For validation, datasets with ground-truth values are used in this paper. Here, we use accuracy, purity, and Normalized Mutual Information (NMI) to evaluate and compare the performances of different clustering methods. All metrics are in the range $[0,1]$, where values close to zero indicate poor clustering and vice versa.

\noindent\textbf{Accuracy ($a$):} We find a permutation ($perm$) of a set of unique values from the resulting labels and find the set that best matches the ground-truth labels. Formally, accuracy ($a$) is defined as $a = \max_{perm \in P} \frac{1}{N} \sum_{i=0}^N V(c_i=d_i)$, where $P$ is the set of all possible permutations of the set of labels acquired from clustering, $V$ is the binary function that yields $1$ when two values are the same and zero for other cases, $c_i \in perm$ is the label for the $i^{\text{th}}$ data point, and $d_i \in G$ is the ground truth label for the $i^{\text{th}}$ data point\,\cite{CoClust}.

\noindent\textbf{Purity ($p$)}: The points in each cluster are all assigned to the ground-truth label, which is most frequent in the cluster. The purity is computed by counting the number of correctly labeled points divided by $N$, the total number of points. Formally, it is defined as $p = \frac{1}{N}\sum_{i\in k}\max_{j}|C_i\cap D_j|$, where $k$ is the number of clusters, $C_i$ is the $i^{\text{th}}$ cluster, and $D_j$ is the $j^{\text{th}}$ ground-truth class\,\cite{purity:original}. $p$ increases with the number of clusters\,\cite{manning}. 

\noindent\textbf{Normalized Mutual Information ($NMI$)}: This metric is defined as $NMI = \frac{2I}{[H(L) + H(G)]}$, where $I$ is the mutual information between $C$ and $G$ and $H(\cdot)$ denotes the entropy\,\cite{nmi:original}.

SDR is performed using \cite{sdr} and all clustering methods and metrics (except NMI) from this paper are implemented in MATLAB. For NMI, we use the implementation by Chen\,\cite{NMI:MATLAB}. We run the experiments on a PC having a Dual-Core Intel Core i5 (2.9 GHz) processor with 8G RAM. 

\section{Results}
\label{sec:results}

\begin{figure}[t!]
  \centering
  \includegraphics[width=0.9\linewidth]{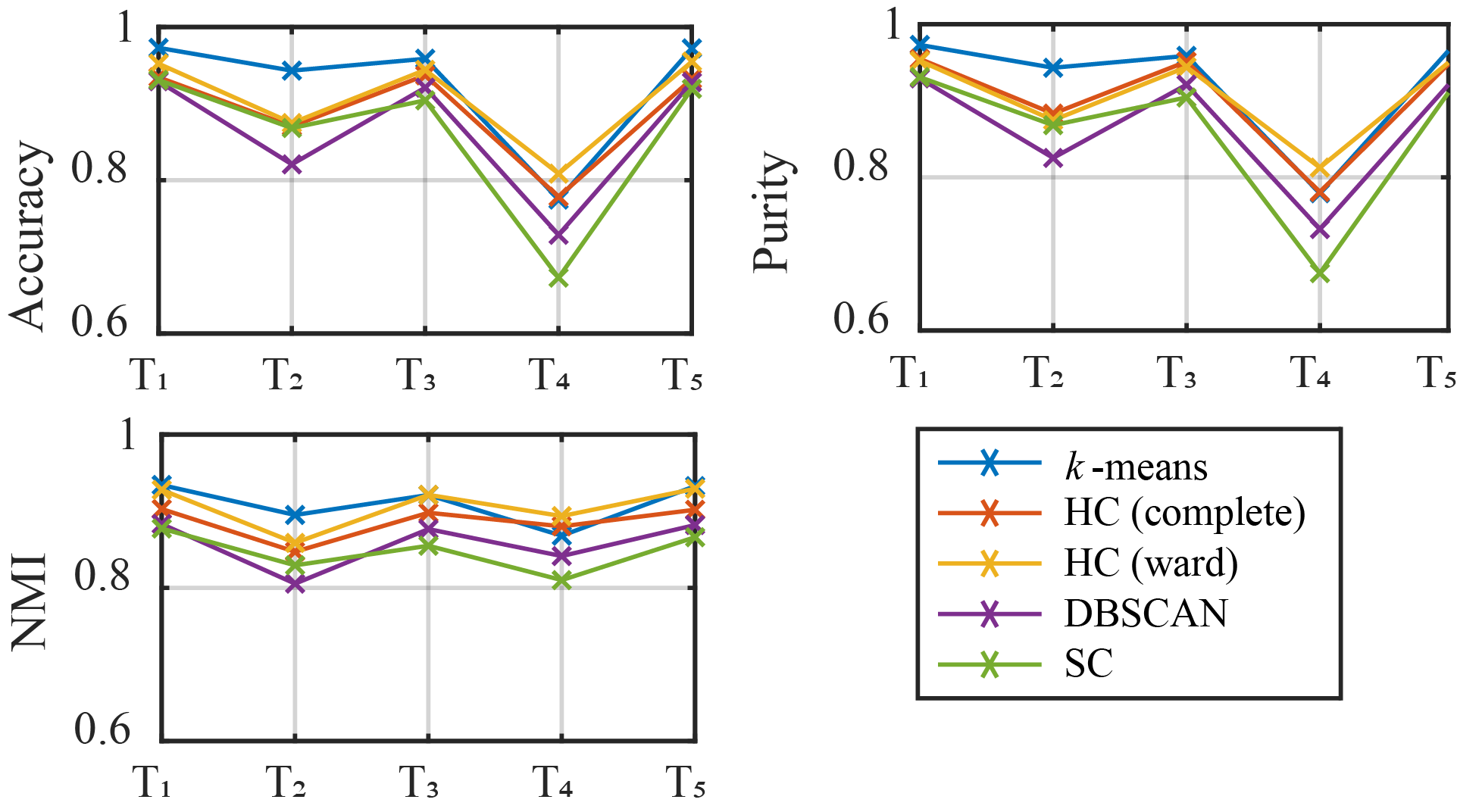}
  \parbox[t]{1\columnwidth}{\relax}
  \caption{\label{fig:syn} Accuracy, purity, and NMI calculated for \emph{k}-means, HC (complete and ward), DBSCAN, and SC results applied to different SLMDS-processed data: equally distributed ($T_1$), varying density ($T_2$), skewed ($T_3$), sub-clustered ($T_4$), and noise ($T_5$) data.
  }
\end{figure}

\subsection{Synthetic data}
\label{sec:results:synData}
We first test our pipeline on the same type of synthetic datasets used in Kim \emph{et al.}\,\cite{sdr} and further aim to replicate the results of \cite{sdr} to compare the clustering performance for DR and SDR. We generate several Gaussian random data sets consisting of $N=5K$ and $n=20$ to cover five types of inter-sample distance distributions: ($T_1$) an even spread of equal Gaussian variance (equal distribution); ($T_2$) an even spread of clusters with different densities; ($T_3$) an uneven spread of clusters (skewed); ($T_4$) sub-clustered data with two pairs of sub-clusters and a one single cluster; ($T_5$) noise (signal-to-noise ratio, $SNR=10$) added to $T_1$\,\cite{sdr}.

We successfully replicated the results from Kim \emph{et al.} and next calculate the evaluation metrics. In Fig.~\ref{fig:syn}, all five methods show high metric values for all types of synthetic datasets except for $T_4$. This is expected because of the low performance of SDR in separating sub-clusters. We also see that $k$-means has the best score of accuracy, purity, and NMI for $T_2$ and $T_3$. We can also observe that $k$-means maintains scores above 0.90 for $T_1$--$T_5$, while the other four methods display a drop in $T_2$. DBSCAN and SC result in the lowest metric scores and SC performs especially poorly compared to the others for $T_1$ and $T_4$.

\begin{figure}[t!]
  \centering
  \includegraphics[width=0.9\linewidth]{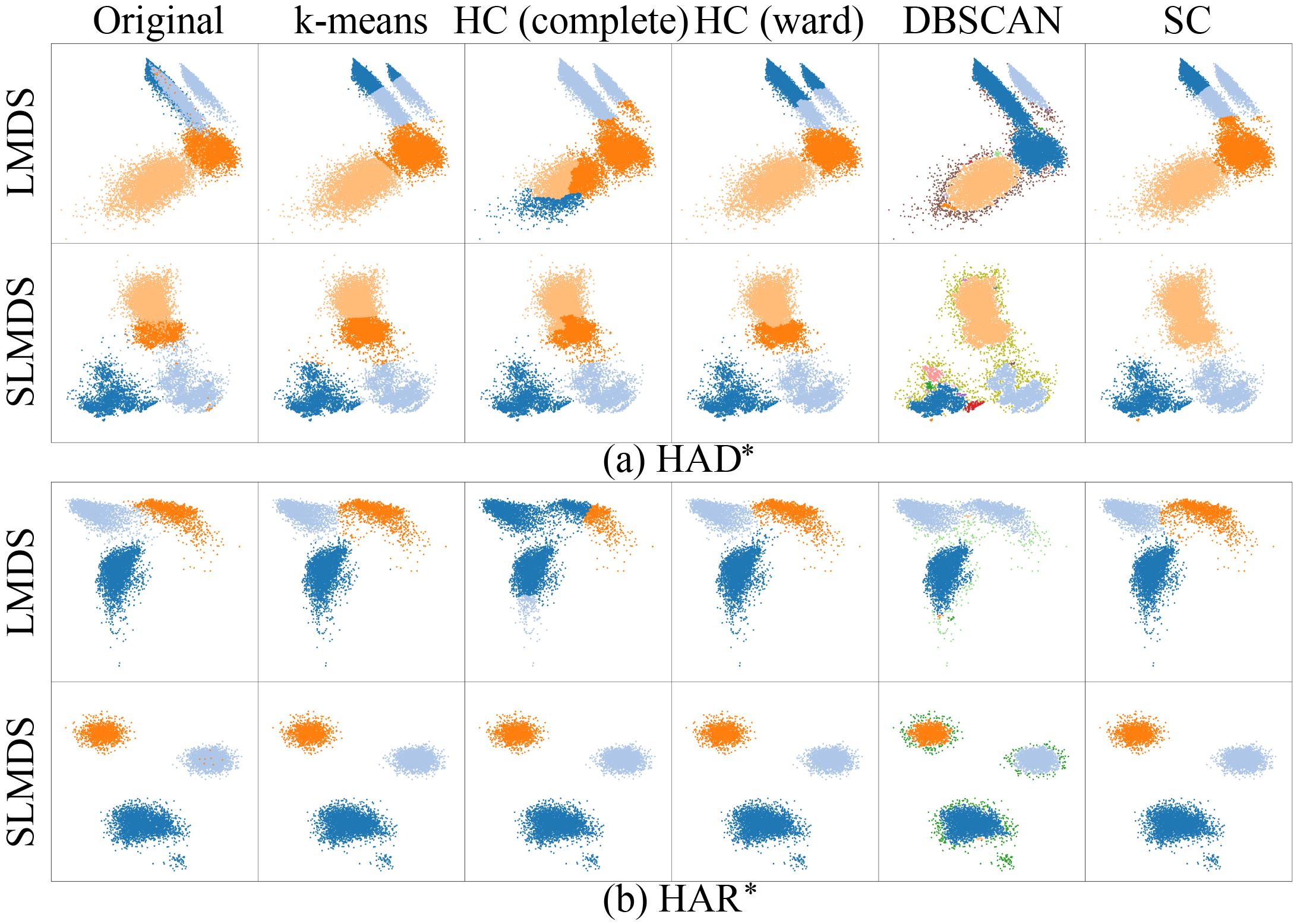}
  \parbox[t]{1\columnwidth}{\relax}
  \caption{\label{fig:human-regroup} Clusters obtained from \emph{k}-means, hierarchical clustering (complete and ward), DBSCAN, and spectral clustering on the LMDS- and SLMDS-processed versions of two human activity data: regrouped human activity data ($n=50$, $N=24075$) and regrouped human activity recognition data ($n=10$, $N=7352$)..
  }
\end{figure}
\begin{table}[t!]
\caption{Average accuracy for \textit{k}-means, HC (complete and ward), DBSCAN, and SC on LMDS- and SLMDS-processed real-world data.}
\label{tab:accuracy}
\centering
\resizebox{\columnwidth}{!}{%
\begin{tabular}{ccccccc}
\hline
                          &       & $k$-means & \begin{tabular}[c]{@{}c@{}}HC\\ (complete)\end{tabular} & \begin{tabular}[c]{@{}c@{}}HC\\ (ward)\end{tabular} & DBSCAN & SC \\ \hline
\multirow{2}{*}{HAD*}     & LMDS  & 0.8149    & 0.5745                                                            & 0.7796                                                        & 0.5730  & 0.8278   \\
                          & SLMDS & 0.9713    & 0.9388                                                            & 0.9637                                                        & 0.7063 & 0.7583   \\ \hline
\multirow{2}{*}{HAR*}     & LMDS  & 0.9829    & 0.4971                                                            & 0.9845                                                        & 0.8107 & 0.9763   \\
                          & SLMDS & 0.9961    & 0.9961                                                            & 0.9961                                                        & 0.9942 & 0.9961   \\ \hline
\multirow{2}{*}{WiFi}     & LMDS  & 0.9160     & 0.8915                                                            & 0.9040                                                         & 0.2714 & 0.9055   \\
                          & SLMDS & 0.9395    & 0.9395                                                            & 0.9395                                                        & 0.9273 & 0.7709   \\ \hline
\multirow{2}{*}{Banknote} & LMDS  & 0.5758    & 0.6793                                                            & 0.5328                                                        & 0.4932 & 0.5364   \\
                          & SLMDS & 0.6844    & 0.7828                                                            & 0.6844                                                        & 0.3134 & 0.5871   \\ \hline
\end{tabular}
}
\end{table}

\subsection{Real-world data}
\label{sec:results:realData}
We use four different SLMDS-processed real-world datasets also tested by Kim \emph{et al.}\,\cite{sdr}.

\noindent\textbf{Human Activity Data (HAD)} consists of human activities recorded using an accelerometer from a smartphone ($N=24075$, $n=60$)\,\cite{uci, realworld:humanActivity, realworld:humanActivity2:calibration, realworld:humanActivity3:feature}. The HAD dataset has five classes (sitting, standing, walking, running, and dancing) and the HAD$^*$ dataset has four super-classes--sitting, standing, normal (walking), and dynamic (running and dancing) movements--defined later.

\noindent\textbf{Human Activity Recognition (HAR)} is from the UCI Machine Learning Repository and contains records of six activities captured using a smartphone\,\cite{realworld:HAR:UCI,uci}. We reduce $n$ from 561 to 10 using principal component analysis (PCA), keeping 80\% of total variance ($N=7352$, $n=10$)\,\cite{sdr}. Activities include lying, standing, sitting, and three walking motions (walking, walking downstairs, and walking upstairs), and HAR$^*$ is defined with three super-classes--lying, sitting or standing, and walking (walking, walking up or downstairs) movements--later defined.

\noindent\textbf{WiFi} is from the UCI Machine Learning Repository and consists of WiFi signal intensities from various routers measured by a smartphone at four different indoor locations ($N=2K$, $n=6$)\,\cite{uci,bhatt_2017, rohra}.

\noindent\textbf{Banknote} is from the UCI Machine Learning Repository\,\cite{uci} and consists of four features extracted using the Wavelet Transform from $N = 1327$ grayscale images of banknote specimens ($n=4$). Samples are labeled as either genuine or forged.

\begin{table}[t!]
\caption{Average purity for \textit{k}-means, HC (complete and ward), DBSCAN, and SC on LMDS- and SLMDS-processed real-world data.}
\label{tab:purity}
\centering
\resizebox{\columnwidth}{!}{%
\begin{tabular}{ccccccc}
\hline
                          &       & $k$-means & \begin{tabular}[c]{@{}c@{}}HC\\ (complete)\end{tabular} & \begin{tabular}[c]{@{}c@{}}HC\\ (ward)\end{tabular} & DBSCAN & SC \\ \hline
\multirow{2}{*}{HAD*}     & LMDS  & 0.8149    & 0.6190                                                             & 0.7796                                                        & 0.5785 & 0.8278   \\
                          & SLMDS & 0.9713    & 0.9388                                                            & 0.9637                                                        & 0.7646 & 0.7672   \\ \hline
\multirow{2}{*}{HAR*}     & LMDS  & 0.9829    & 0.5257                                                            & 0.9845                                                        & 0.8139 & 0.9763   \\
                          & SLMDS & 0.9961    & 0.9961                                                            & 0.9961                                                        & 0.9959 & 0.9961   \\ \hline
\multirow{2}{*}{WiFi}     & LMDS  & 0.9160     & 0.8915                                                            & 0.9040                                                         & 0.2781 & 0.9055   \\
                          & SLMDS & 0.9395    & 0.9395                                                            & 0.9395                                                        & 0.9373 & 0.7825   \\ \hline
\multirow{2}{*}{Banknote} & LMDS  & 0.5758    & 0.6793                                                            & 0.5554                                                        & 0.7252 & 0.5554   \\
                          & SLMDS & 0.6844    & 0.7828                                                            & 0.6844                                                        & 0.9785 & 0.5906   \\ \hline
\end{tabular}
}
\end{table}
\begin{table}[t!]
\caption{Average NMI for \textit{k}-means, HC (complete and ward), DBSCAN, and SC on LMDS- and SLMDS-processed real-world data.}
\label{tab:nmi}
\centering
\resizebox{\columnwidth}{!}{%
\begin{tabular}{ccccccc}
\hline
                          &       & $k$-means & \begin{tabular}[c]{@{}c@{}}HC\\ (complete)\end{tabular} & \begin{tabular}[c]{@{}c@{}}HC\\ (ward)\end{tabular} & DBSCAN & SC \\ \hline
\multirow{2}{*}{HAD*}     & LMDS  & 0.7101    & 0.5930                                                             & 0.7471                                                        & 0.5880  & 0.7077   \\
                          & SLMDS & 0.9090     & 0.8441                                                            & 0.8969                                                        & 0.7656 & 0.8106   \\ \hline
\multirow{2}{*}{HAR*}     & LMDS  & 0.9306    & 0.2566                                                            & 0.9313                                                        & 0.7975 & 0.9110    \\
                          & SLMDS & 0.9767    & 0.9767                                                            & 0.9767                                                        & 0.9705 & 0.9767   \\ \hline
\multirow{2}{*}{WiFi}     & LMDS  & 0.7887    & 0.7386                                                            & 0.7611                                                        & 0.0549 & 0.7759   \\
                          & SLMDS & 0.8686    & 0.8686                                                            & 0.8686                                                        & 0.8536 & 0.7682   \\ \hline
\multirow{2}{*}{Banknote} & LMDS  & 0.0184    & 0.2068                                                            & 0.0045                                                        & 0.2106 & 0.0123   \\
                          & SLMDS & 0.2094    & 0.4044                                                            & 0.2094                                                        & 0.5311 & 0.0718   \\ \hline
\end{tabular}
}
\end{table}

\subsection{Qualitative evaluation}
\label{sec:results:qualEval}
We show in Fig.~\ref{fig:human-regroup} the projections color-coded based on their clustering results of HAD$^*$ and HAR$^*$ datasets. Note that we here mainly show results with super-class labels (HAD$^*$ and HAR$^*$) and add the results of sub-class labels (HAD and HAR) to the supplemental materials (\cite{jeewonHeo}) instead. We discuss these later in Sec.~\ref{sec:discussion}\hspace{0.3cm}. For the  HAD$^*$ dataset, Fig.~\ref{fig:human-regroup} shows that the cluster related to walking is placed close to the cluster dynamic movements, and the other two clusters for sitting and standing have more distance between each other and from other activities. Overall, $k$-means and HC perform better at separating the four super-classes compared to other clustering methods. Further note that the clustering performances (i.e., accuracy, purity, and NMI) of most clustering methods are higher for SLMDS-processed data than LMDS-processed data due to the sharpening effect.

For the HAR$^*$ dataset, the LMDS projection displays three visually distinguishable clusters, which are placed very close to each other and we observe a lot of noise near the boundaries of the clusters. This makes the clustering more challenging near the boundaries. This phenomenon is most evident in HC (complete) and DBSCAN. In contrast, SLMDS produces three well-separated spherical groups. All five clustering methods seem to correctly separate the data into three distinct clusters. Due to space limitations, clustering results for WiFi and Banknote datasets are added to the supplemental materials (\cite{jeewonHeo}).

\subsection{Quantitative evaluation}
\label{sec:results:quantEval}
We compare next the evaluation metrics introduced in Sec.~\ref{sec:method}\hspace{0.3cm}\ref{sec:method:eval}.\hspace{0.2cm} Tables~\ref{tab:accuracy}--\ref{tab:nmi} each show different evaluation scores for real-world datasets. Overall, we observe from the clustering evaluation metrics that $k$-means performs the best and SC performs the worst (lower accuracy and purity scores and little difference in NMI). HC (complete and ward) shows similar scores, but HC (ward) produces slightly higher scores than HC (complete) for the HAD$^*$ and HAR$^*$ datasets.

Tables~\ref{tab:accuracy}--\ref{tab:nmi} show that accuracy, purity, and NMI scores for SLMDS are higher than those for LMDS using all real-world datasets, except the Banknote dataset (see Sec.~\ref{sec:discussion}\hspace{0.3cm} for more discussion and limitations). We observe further that SLMDS improves all three evaluation metric scores for most clustering methods ($k$-means and HC complete and ward). Especially for the WiFi and HAR$^*$ datasets, the scores were near unity (highest score).

\section{Discussion}
\label{sec:discussion}
\noindent\textbf{Selection of clustering methods}
The selection is based on the method's availability, ease-of-use, and performance. To show the overall performance of the selected clustering methods regardless of SLMDS and LMDS, we computed the quantitative metrics for plain 2D synthetic data (varying the number of clusters and $N$). We used ten datasets for each combination to calculate the average performance of each clustering method. Results show that the three quantitative metrics for all five clustering methods are above $0.9$ for all datasets when the number of clusters are set to five (see supplemental materials (\cite{jeewonHeo})).

\noindent\textbf{Computational scalability}
The time complexity of \emph{k}-means is $O(I\,k\,N\,n)$, where $I$ is the number of iterations, $k$ is the number of clusters, $N$ is the number of observations, and $n$ is the number of dimensions\,\cite{manning}. The time complexity is $O(N^3)$ for HC\,\cite{day}, $O(N\log N)$ for DBSCAN\,\cite{dbscan:original}, and $O(N^3)$ for SC\,\cite{Tsironis}. We computed the wall-clock time measurements using the same dataset from above and conclude that $k$-means clustering is the fastest among the five clustering methods (see supplemental materials\,\cite{jeewonHeo} for detailed results).
\begin{figure}[t!]
  \centering
  \includegraphics[width=0.9\linewidth]{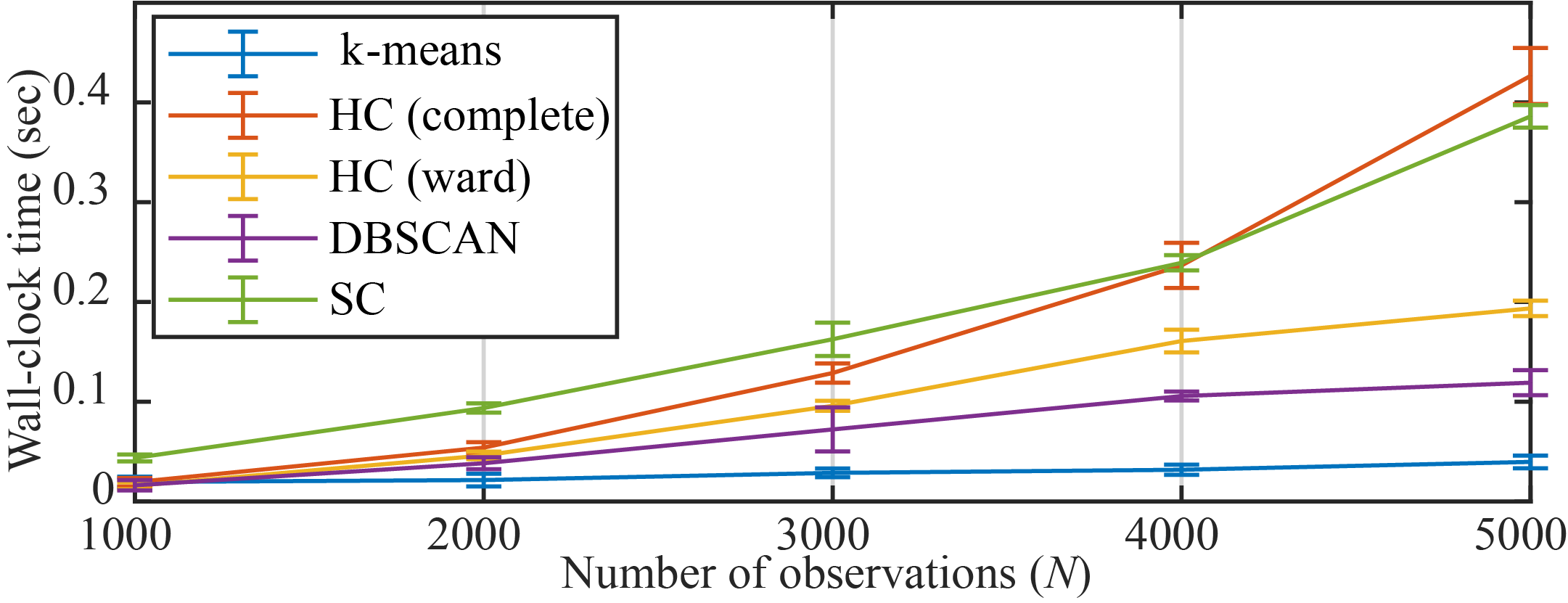}
  \parbox[t]{1\columnwidth}{\relax}
  \caption{\label{fig:time}Wall-clock timing measurements of different clustering methods using varying number of observations ($N$) in 2D.}
\end{figure}

\noindent\textbf{Limitations}
Applying clustering to SLMDS has some limitations. First, the parameter settings for both clustering and SLMDS are not entirely automatic, which requires time and effort from the end-users to fully explore the parameter space. Next, clustering sub-clusters is still challenging. SLMDS by nature captures the super-class structures better than sub-classes (refer to clustering metric scores of synthetic datasets in the supplemental materials\,\cite{jeewonHeo})\,\cite{sdr}. This means that the performance of clustering largely depends on SLMDS and that in some cases clustering will perform badly even though clusters are well-separated as in Fig.~\ref{fig:human-regroup}.

\section{Conclusion}
\label{sec:conclusion}
We presented a pipeline consisting of SDR and clustering to ease the manual labeling process for end-users using SDR. We tested our pipeline using both synthetic and real-world datasets and compared the clustering performances of SDR- and DR-processed data. Overall, the qualitative and quantitative results verified that the highly separated clusters produced by SDR yield higher clustering performance (than DR), which has not been shown before. Furthermore, we recommend $k$-means for clustering human activity data based on the ease-of-use, performance, and computational scalability. This proposed pipeline and set of validation metrics will ultimately provide a user-friendly environment for data exploration and semi-automatic classification.

\section*{Conflict of Interest}
The authors declare that they have no conflict of interest.

\section*{Acknowledgements}
This work is supported by DSSC Doctoral Training Program co-funded by the Marie Sklodowska-Curie COFUND project (DSSC 754315).

\bibliography{cites}

\begin{table}[b!]
\footnotesize 
        \begin{tabular}{ll}
        &Author: Youngjoo Kim\\
        &Institute: Bernoulli Institute, University of Groningen\\
        &Street: Nijenborgh 9 (9747AG)\\
        &City: Groningen\\
        &Country: The Netherlands\\
        &Email: lyoungjookiml@gmail.com\\
        \end{tabular}
\end{table}

\end{document}